\RequirePackage{fix-cm}
\documentclass[twocolumn]{svjour3} 
\smartqed  
\usepackage{graphicx}
\usepackage{amsmath}
\usepackage[ruled,vlined]{algorithm2e}
\usepackage{makecell}
\usepackage[utf8]{inputenc}
\usepackage{xcolor}
\usepackage{hyperref}
\usepackage{subcaption}
\usepackage{float}
\usepackage{pgfplots}
\pgfplotsset{compat=1.17}
\usepackage{adjustbox}
\begingroup\newif\ifmy
\IfFileExists{plotData.csv}{}{\mytrue}
\ifmy

\fi\endgroup

\begin{document}


\title{Development of a robust cascaded architecture for intelligent robot grasping using limited labelled data
}

\author{Priya Shukla \and Vandana Kushwaha \and G. C. Nandi}

\institute{ Priya Shukla \at
              Student at Center of Intelligent Robotics, Indian Institute of Information Technology Allahabad, Prayagraj-211015, U.P., INDIA \\
              \email{priyashuklalko@gmail.com} 
        \and
          Vandana Kushwaha \at
              Student at Center of Intelligent Robotics, Indian Institute of Information Technology Allahabad, Prayagraj-211015, U.P., INDIA \\            
          \and
         G. C. Nandi \at
              Professor at Center of Intelligent Robotics, Indian Institute of Information Technology Allahabad, Prayagraj-211015, U.P., INDIA \\
              \email{gcnandi@iiita.ac.in}
}

\date{Received: date / Accepted: date}

\maketitle

\section*{List of abbreviations}
VAE: Variational Auto-Encoder\\
VQVAE: Vector Quantized Variational Auto-Encoder\\
CNN: Convolutional Neural Network \\
GGCNN: Generative Grasp CNN \\
GGCNN2: Generative Grasp CNN-2 \\
RGGCNN: Representation based GGCNN \\
RGGCNN2: Representation based GGCNN2 \\

\begin{abstract}
Grasping objects intelligently is a challenging task even for humans and we spend a considerable amount of time during our childhood to learn how to grasp objects correctly. In the case of robots, we can not afford to spend that much time on making it to learn how to grasp objects effectively. Therefore, in the present research we propose an  efficient learning architecture based on VQVAE so that robots can be taught with sufficient data corresponding to correct grasping. However, getting sufficient labelled data is extremely difficult in the robot grasping domain. To help solve this problem, a semi-supervised learning based model which has much more generalization capability even with limited labelled data set, has been investigated. Its performance shows 6\% improvement when compared with existing state-of-the-art models including our earlier model.
During experimentation, It has been observed that our proposed model, RGGCNN2, performs significantly better, both in grasping isolated objects as well as  objects in a cluttered environment, compared to the existing approaches which do not use unlabelled data for generating grasping rectangles. To the best of our knowledge, developing an intelligent robot grasping model (based on semi-supervised learning) trained through representation learning and exploiting the high-quality learning ability of GGCNN2 architecture with the limited number of labelled dataset together with the learned latent embeddings, can be used as a de-facto training method which has been established and also validated in this paper through rigorous hardware experimentations using Baxter (Anukul) research robot (\href{https://youtu.be/_fbuYC0J9Y0}{Video demonstration}).


\end{abstract}

\section{Introduction}
Making a robot capable of manipulating objects in a dynamically changing real world environment skillfully, the way we do, is extremely difficult. It turns out that for us also the object manipulation through grasping is challenging and it requires substantial learning. A child generally has a poor skill of grasping and thus they are normally not allowed to manipulate sophisticated/fragile items. But over the years, through training, kids grow skill and learn how to grasp objects with various shapes and sizes appropriately, making a grown up person capable of handling those items safely. Imparting such skill to a robot, although required, is extremely challenging due to the nonavailability of sufficient training data, since for making a robot capable of learning from experience, quality data are needed. This challenge has been drawing the attention from the researchers worldwide for many years. Robot grasp pose estimation has also been considered as most complex problem not only due to the dynamically changing object manipulation environment (like illumination and pose changes of the objects to be grasped) but also due to the large categories of objects with different texture, shape or size required to be manipulated. 

Previously, rigorous research has focused mainly on grasp planning, considering the different aspects of grasp pose synthesis in \cite{c27,c3} with limited success, due to its lack of learning ability and thus grasp manipulations are confined to known objects only. With the advent of machine learning and deep learning techniques, researchers are trying to make grasping models to  generalize grasp manipulation for unseen/unknown objects with the feature learning from quality robot grasp \cite{lenz2015deep,c21,c23}. Initially, grasp detection has been considered as a computer vision problem which is further modified as a grasp rectangle detection \cite{kumra2017robotic,redmon2015real} problem achieving greater success. Primarily, in these approaches they find all the candidate grasp rectangles using deep neural networks and then further determine the optimal grasp rectangle using candidate rectangles with the shallow neural network. This makes the grasp pose detection more time consuming process.

Models based on all these approaches, although work well in the laboratory environment, fail to work in the real world environment. To overcome this problem and dependencies, pixel wise anthropomorphic grasp prediction has been solved in \cite{agc} but it is limited to regular shaped objects only. To revamp the pixel wise grasp prediction in \cite{morrison2018closing} authors proposed the generative grasp concept which reduces the neural network parameters and makes its execution faster for optimal grasp predictions. To improve the performance, the architecture given in \cite{morrison2018closing} has been further modified in \cite{ggcnn2}. Recently, authors in \cite{VQVAE} propose a discrete representation learning with vector quantization  which works significantly well for image \cite{n38,n12}, audio \cite{n37,n26} and video \cite{n20,n11} datasets.

Owing to the abundance of unlabelled data and scarcity of labelled data, we propose to design a model which can utilize the unlabelled data together with whatever labelled data we have, saving on the expenses of a large labelled dataset. Further, the neural network should have the ability to generalize to the extent of understanding the data and the semantics behind it, rather than learning only the mapping between the input space and the output space.  We have named our model as Representation based  Generative Grasp Convolutional Neural Network 2 (RGGCNN2) which uses VQVAE for representation learning in the encoding space and the GGCNN2 as a decoder for predicting the grasp patch. In doing so we have exploited the variational auto-encoder (VAE) paradigm, a powerful class of probabilistic models that facilitate generation and has the ability to model complex distributions.Due to it's inherent strength,our architecture could produce enough unlabelled data to generalize better in producing optimal grasp rectangles/patches even with limited number of labelled dataset. It works efficiently and perform significantly better compared to the existing models  in grasping objects in an isolated as well as cluttered environment for seen as well as unseen objects. Moreover, all the generated grasp for the tested objects have also been verified in a real world environment with the Anukul (Baxter) research robot. More specifically, following are our major research contributions:
\begin{itemize}
    \item The grasp patches have been generated using a semi supervised learning approach where it uses the VQVAE architecture for identifying the important features of the pixels in an image. Initially, VQVAE model is trained in an unsupervised way using an unlabelled dataset to regress the discrete latent representation. Due to this capability the proposed approach is able to work significantly better  with the limited number of labelled dataset. 
    \item Subsequently, the existing model of VQVAE has been augmented  with the Generative Grasp Convolutional Neural Network 2, (GGCNN2), for utilizing the vector quantized discrete latent space efficiently in grasp patch generation and the generated grasp patches in the image configuration space have been mapped in robot configuration space to execute the grasps successfully with the physical robot(Anukul).
    \item With the proposed modular solution, the robot is able to grasp the object, including novel objects successfully in  isolated as well as cluttered environments.
    \item Developed a simple but elegant  strategy of safe table top grasping for the target object without hitting other objects in the working environment  

\end{itemize}

Rest of the paper has been organized into six major sections as follows: 
Section 1 introduces the problem with insights of our major contributions. Section 2 critically analyses the previous research undertaken in this area. Detailed overview of the proposed approach has been discussed in the Section 3. All the prerequisite concepts with the proposed RGGCNN2 to robot execution have been detailed in Section 4. Section 5 demonstrates the experimental setup and analyses the experimental results with existing and proposed approaches. Finally, the conclusion and future scope of research has been discussed in Section 6.

\section{Analyses of previous related research}
Robot grasping within an unstructured environment for novel object is an immense challenge. Robot grasping has been considered as a computer vision problem because to grasp a novel object with optimal pose involves image based features. This strategy is referred as \textit{Robot for Vision} where based on image features robot calculates its pose to grasp an object successfully. Earlier research are  mostly focused on hand knitted features \cite{he} which is very monotonous and tiring task but required for data driven grasp approaches \cite{c3}. With the advent of deep learning, computer vision has shown excellent improvements in areas like object detection and localization\cite{od1,od2}. 

Predominantly, grasping is a fundamental problem of robotics, which is mutating year after years like planning \cite{c3,c27} to learning \cite{c23,p9}. Due to the availability of giant deep network architectures coupled with huge data processing power, deep learning is playing a very significant role in grasp prediction for seen and unseen objects \cite{d1,d11}. Grasp detection with grasp patch (grasp rectangle) estimation has brought revolutionary improvement in deep learning based robot grasping \cite{redmon2015real,13,14,lenz2015deep,wei2017robotic,wang1999lagrangian}.

Most of such deep learning based grasping models follow two fold architectures, first the candidate grasp rectangles are sampled from the input image and then they are fed to the Convolutional Neural Network (CNN) to find the optimal grasp rectangle from the sampled grasp rectangles. This makes the computation exhaustive which leads the grasp execution process suitable only for an open loop grasping. Open loop grasping does not include feedback from the working environment causing grasp execution vulnerability to any changes of location and orientation of the objects causing grasp failure. 
Few works have also been directed towards regressing the optimal grasp rectangle with a single neural network as presented in \cite{redmon2015real,kumra2017robotic}. However, regressing an optimal grasp rectangle for the input image might be averaging all the possible grasp rectangles for the target object. Sometimes, it cannot be a relevant optimal grasp candidate because  normal averaging may not include the spatial features of the targeted objects of an image. 

In \cite{morrison2018closing}, authors  introduce the Generative Grasp Convolution Neural Network (GGCNN) which predicts pixel wise grasp patch in a dynamic environment. Moreover, in this research authors reduce grasp patch prediction time which is very appealing for the closed loop grasping. In closed loop grasping, if any changes occur in object pose after grasp patch prediction, it takes feedback from the environment and try to predict grasp patch again for the updated pose of the object. GGCNN architecture has been modified in \cite{ggcnn2} by introducing dilated convolutional layers in GGCNN architecture and presented as GGCNN2 for the improved performance. 
Somehow the results of such approaches with better generalization largely depend on availability of the huge labelled dataset which is extremely difficult to obtain. Although some labelled dataset like Cornell grasping dataset and few such \cite{cgd,jacquard} are available, they are confined to some limited objects and able to generate  optimal grasp detection with grasp patch only on those objects or some closely related objects which the network has seen during training.
Grasp patch generation for the novel objects, we require huge labelled data for training which is in scarcity and the dream of accomplishing intelligent grasping, the way we grasp, remains unfulfilled. To overcome this limitation, in the present investigation a novel idea of introducing generative models to produce grasp patch generations even for novel objects has been proposed. To the best of our knowledge, using generative model for learning latent embeddings and generating grasp patch in a semi supervised way using whatever labelled data available together with the learned latent embedding, is a new concept. The details about the proposed approach have been discussed in the subsequent sections.   

\section{Proposed approach}

\begin{figure}
    \centering
    \includegraphics[scale=0.5]{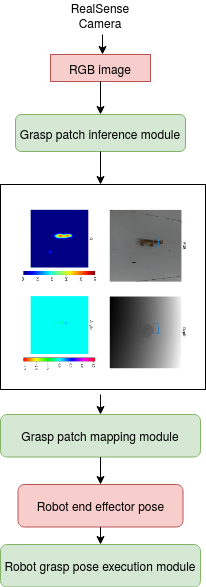}
    \caption{Grasp Patch execution}
    \label{flow}
\end{figure}

In the proposed approach, complete robot grasping, from training to execution  has been presented in Fig. \ref{flow}. The grasp patch inference module takes images as an input and predicts the grasp patch with Width (W), Quality (Q) map and Angle map using the trained RGGCNN2 model. The detailed procedure of the training RGGCNN2 model has been elaborated in Section \ref{sec:RGCNN2}. 

Subsequently, mapping between image grasp patch pose and End Effector (EE) pose has been done with the help of grasp patch mapping module. Mapping details have been discussed in Section \ref{sec:gpm}.  
Afterwards, robot grasp pose execution module calculates the joint angles and controls the arm movement with the help of inverse kinematics joint solutions and trajectory planning.
Explanation for the robot grasp pose execution module has been clearly described in Section \ref{sec:gpe}.

With the help of the above mentioned modules, robot is able to grasp any object with the predicted grasp patch in isolated as well as cluttered environment. The developmental details of the proposed approaches have been discussed next in the Methodology section. 

\section{Methodology}
\subsection{Preliminaries}
\subsubsection{Representation learning models}
Explicit modelling distribution for the unlabelled images or data exhibit the latent variables which help in supervised learning of grasp patches. The VAE approximates the density estimation for the training data where density function has been defined in (\ref{eqn:df}) where z, x denotes the set of latent variables and the set of observed variables respectively.
\begin{equation} \label{eqn:df}
p_\theta(x) = \int_{z} p_\theta(z)p_\theta(x|z)dz
\end{equation} 

We can see that density function is intractable due to the intractability of the posterior distribution \cite{PD}. VAE uses two probabilistic models such as encoder model and decoder model. Encoder model is used for inference whereas decoder model is used for generation. The objective function of VAE is to maximize the Evidence Lower BOund (ELBO). Basically, ELBO is a tractable lower bound, which can be optimized by calculating its gradient using reparameterization trick \cite{RPT}.

\begin{equation}\label{eqn:elbo}
ELBO = E_z\big[\log p_\theta(x|z)\big]  -  D_{KL}(q_\phi(z|x)||p_\theta(z))
\end{equation} 
In (\ref{eqn:elbo}), the first part amplifies the log likelihood whereas second part estimates the Kullback Leibler (KL) divergence \cite{KL} between the  approximate posterior and the true posterior. It has a closed form solution for the vanilla VAE model in which both the distributions are assumed to be Gaussian. Encoder model has the ability to infer the $q_{\phi}(z|x)$ which can be used for representation based learning.

VAE model consists discrete and continuous latent variables. In \cite{st}, the authors suggest that the images can be modelled better with the discrete latent variable only. The situation when VAE model’s output doesn't depend on the latent variables due to excessive training of the decoder is called as posterior collapse. To circumvent the posterior collapse, the authors presented a Vector Quantized Variational AutoEncoder (VQVAE) \cite{VQVAE} which uses only discrete latent variable instead of continuous latent variable.


\begin{figure*}
	\centering
		\includegraphics[scale=.3]{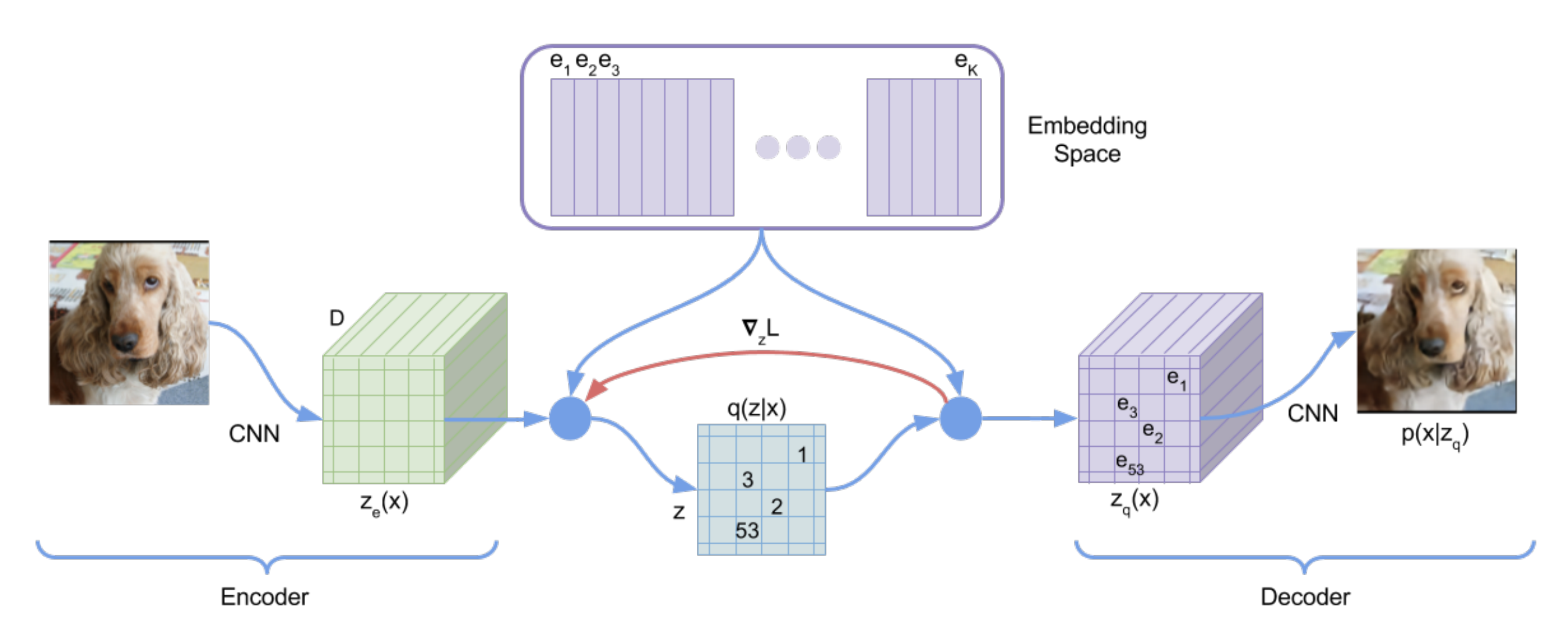}
    	    \caption{Vector Quantized Variational AutoEncoder \cite{VQVAE}}
	    \label{fig:VQVAE}
\end{figure*}
In VQVAE, encoder inputs an image and renders continuous-valued vector ${z_e(x)}$ as an output. Subsequently, it has been utilized by the Vector Quantizer (VQ) for accomplishing a nearest neighbour search for embeddings in the dictionary. The flow of VQVAE model has been presented in Fig. \ref{fig:VQVAE}. For all embedding ${e \in {R}^{D \times K}}$,  with embedding dimensionality D, K represents the total count of embeddings in the embedding space. In essence, VQ, quantifies the encoder's output and yields ${z_q(x)}$, finally  forwarded for attaining the reconstruction from the decoder. The objective of training has been shown in (\ref{eqn:L}) where, sg denotes the stop gradient operation:

\begin{equation}\label{eqn:L}
    L = \text{log } p(x|z_q(x)) + ||\text{sg}[z_e(x)]-e||_{2} ^2\  +\  \beta||z_e(x)-\text{sg}[e]||_{2} ^2
\end{equation} 

Remember that a prior through the embeddings of dictionary follow a uniform distribution. The foremost term, portrays the outcome of variational inference performed for acquiring the ELBO, the middle or dictionary-learning term, shifts the latent embeddings in close proximity with the encoder's output, and the last term signifies the commitment-loss, that shifts the encoder's output towards the specified embedding selected from the dictionary.

\subsubsection{Generative grasp patch generation models}
Generative grasp patch generation models have been given by \cite{morrison2018closing,ggcnn2} which is termed as GGCNN and GGCNN2. It abolishes the two stage grasp detection pipeline and reduces the time complexity by minimizing the network parameters. Both the mentioned models predict pixel wise  grasp for each pixel in image (I). It takes depth image as an input and predicts grasp parameters $g=[p,\phi,w,q]$ where p is for pixel coordinate (u,v) which is being center of the grasp patch, $\phi$ is for the rotation angle or orientation, w is opening width of gripper and q is for quality measure of the grasp patch. To achieve the function M which has been approximated by network models with image I, having dimension H as height and W as width.
\begin{equation}\label{eqn:gg}
    G^{3\times H\times W}=M(I^{H\times W})
\end{equation}
G denotes the grasp map which defines the grasp set for all image pixels, with all three parameters $(\Phi,W,Q)^{3\times H\times W}$ where optimal grasp is estimated by its maximum quality (q) grasp parameters which ranges from 0 to 1.

\begin{figure*}
    \centering
    \includegraphics[scale=0.6]{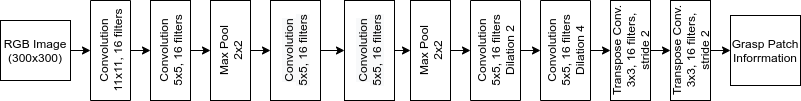}
    \caption{GGCNN2}
    \label{fig:GGCNN2}
\end{figure*}

The architecture of GGCNN contains only convolution and transposed convolution layers with distinct dimensions of filter and stride whereas, in GGCNN2 architecture some additional dilated convolutional layers are added to improve the performance. The detailed architecture of GGCNN2 has been shown in Fig. \ref{fig:GGCNN2}. 
Both the above mentioned models uses same RGB images as an input and predicts the grasp patch in our experiments. Both the models have been trained using our Central computing facility where total estimated parameters have been shown to be 67,604 and 70,548 for GGCNN and GGCNN2 respectively.

\begin{figure*}[t]
	\centering
		\includegraphics[scale=.5]{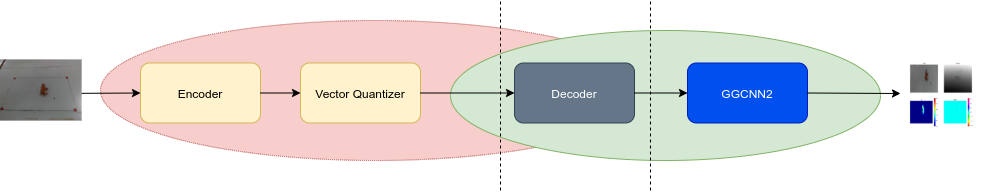}
	    \caption{Representation based  Generative Grasp Convolutional Neural Network 2 (RGGCNN2)}
	    \label{fig:rggcnn2}
\end{figure*}
\subsection{Proposed model (RGGCNN2)}
\label{sec:RGCNN2}
In this paper, semi supervised grasp patch generation approach has been proposed. For training, dataset is distributed into two different parts such as labelled data (n1) and unlabelled data (n2). Labelled data contains corresponding grasp vectors for each RGB image. 
It is expressed as a pair, $(x,y)$, where $x$ represents an RGB-image while $y$ represents grasp vectors corresponding to that RGB-image. Contrarily, unlabelled data solely contains RGB images. The proposed model embodies an encoder, a vector-quantizer, a decoder, accompanied by the GGCNN2 layers. Primarily, the encoder, the vector-quantizer, and the decoder layers are trained on all the RGB images of labelled and unlabelled training samples. Predominantly, the first phase of training, indicates the training of a VQVAE.
Subsequently, the encoder and the vector quantization layer weights are freezed. Concretely, this fixes the output of the vector quantizer as the embedding chosen by the VQVAE trained earlier. Thereafter, we reinitialize the decoder weights. This can be procured as enhancing the GGCNN2 network with the decoder. The certainty of our conclusion is based on the fact that the decoder efficiently utilizes the embeddings generated by the vector-quantizer. The predicted GGCNN2 grasp value, ${\tilde{g}}$,  is expressed as ${\tilde{g}} = (\tilde{p},\tilde{w}, \tilde{\phi}, q)$, where $\tilde{p}$, $\tilde{w}$, and $\tilde{\phi}$ denote the position $(i,j)$ of pixel in an image corresponds to the center of grasp patch/rectangle, opening width and rotation angle of the gripper, respectively. The GGCNN2 yields three maps for the identical dimension of the inputted image, that is, the output is in the form of $({W}, {\Phi}, {Q})^{3 \times H \times W}$, where W, ${\Phi}$ and Q provide pixel-wise values for $\tilde{w}$, $\tilde{\phi}$ and $\tilde{q}$, respectively. The optimal grasp rectangle is realized using the pixel conforming to the highest $\tilde{q}$ value. Finally, we train the thorough model on the labelled data only. The complete training procedure of our proposed model, RGGCNN2, is summarized in Fig. \ref{fig:rggcnn2}.

\subsection{Grasp patch mapping and evaluation}
\subsubsection{Grasp patch mapping}
\label{sec:gpm}
\begin{figure}
	\centering
		\includegraphics[scale=.55]{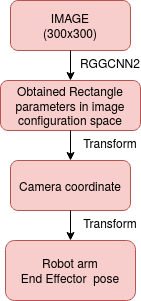}
	    \caption{Grasp patch mapping}
	    \label{fig:Gmap}
\end{figure}

After obtaining grasp patches in the image configuration space from the above mentioned trained model, it is mapped with the EE pose to execute the robotic grasp in a real world environment. Entire grasp patch mapping is divided into two major parts. Primarily, image coordinates needs to map/transform into camera coordinates using camera parameters. Secondly, camera coordinates have been mapped with the EE pose. The complete grasp patch mapping has been represented in Fig. \ref{fig:Gmap}.
As shown in the figure, from the graspable object center $\tilde{p}(i,j) $, obtained from RGGCNN2 in the image plane, we can directly get the position of the object with respect to the camera. Orientation of the same object as obtained from the same RGGCNN2 has also been calculated with respect to the camera using inverse  Euler angle, Roll-Pitch-Yaw, for which the obtained values are $\pi$, 0 and $\theta$ respectively. For experiments Anukul (Baxter) robot has been used. It has seven degrees of freedom (DoF) in each arm. In this research left arm has been used, however, without loss of generality, the right arm can also be used. Robot arm EE pose consists seven parameters where three are for a position in X-Y-Z planes and four are for a orientation using the quaternion representation. 
\begin{figure}
	\centering
		\includegraphics[scale=.40]{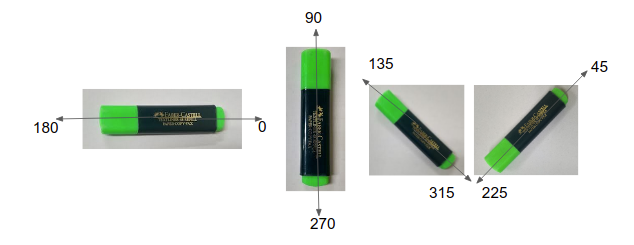}
	    \caption{Orientation observations for each position}
	    \label{fig:pmap}
\end{figure}
For orientation transformation we have used the Robot Operating System (ROS) transformation (tf) library to convert the Euler angle representation to quaternion angle representation. Robot mapping with the camera is represented in (\ref{eqn:RC}) where R is robot matrix and C is the Image matrix. For mapping we have taken 10 object positions with their four different orientations. Positions have been considered within the defined robot workspace and for orientation, we have considered the object placing angles such as 0$^{\circ}$, 45$^{\circ}$, 90$^{\circ}$, 135$^{\circ}$. Here, we have assumed object orientation with $\Theta$ is same as $\Theta+\Pi$ because both makes the object pose in the identical alignment. So object orientation has been restricted to only four $\Theta$ values.  In our mapping the total number of observations is 40 which is the product of 10 positions  and 4 orientations for each position (as shown in Fig. \ref{fig:pmap}). 

\begin{equation}\label{eqn:RC}
    R=T_m(C)
\end{equation}
As evident, the dimensions of R matrix and the C matrix depend on the number of observations. For each observation the dimension of R is 7$\times$1 and that of C is also 7$\times$1. For n number of observations, the dimension of C becomes 7$\times$n and to obtain the mapping matrix $T_m$, we need to calculate the pseudoinverse of C and when this pseudoinverse is multiplied by R generates a 7$\times$7 mapping matrix $T_m$. Hereafter, for any pose C, we will be able to get a corresponding R (robot EE pose) which is executed, as described in the following section.
\begin{figure*}[!t]
	\centering
		\includegraphics[scale=.70]{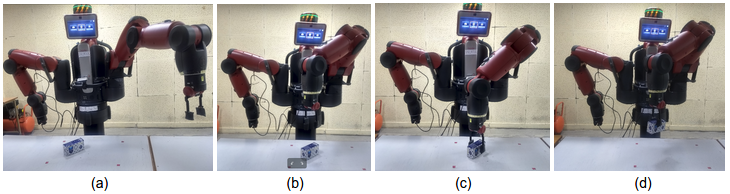}
	    \caption{Grasp evaluation phases}
	    \label{fig:rexec}
\end{figure*}

\subsubsection{Grasp patch evaluation}
\label{sec:gpe}
In the grasp patch evaluation, robot grasp pose execution module calculates the joint angles for the specified pose of the robot arm EE. Joint angles for a pose have been calculated using inverse kinematics \cite{BK}, which has been shown by (\ref{eqn:Ik}). 
\begin{equation}\label{eqn:Ik}
 (\theta\textsubscript{1},\theta\textsubscript{2},\theta\textsubscript{3},\theta\textsubscript{4},\theta\textsubscript{5},\theta\textsubscript{6},\theta\textsubscript{7})= IK(POSE)   
\end{equation}
To execute a grasp in the real world, we have designed a simple but elegant method of inverse kinematics solution and trajectory planning for the arm movement within its working envelope which has been shown as a robot grasp pose execution module in the schematic diagram. One complete cycle of grasp evaluation has been shown in Fig. \ref{fig:rexec}. There are four steps as described below:
\begin{itemize}
    \item Firstly, the EE pose is initialized as neutral pose (home position of Anukul).
    \item Next, to avoid collision with the surrounding objects, the trajectory has been planned in such a way that during movement, the z component (in X-Y-Z plane) for the position of the gripper remains at a sufficient height (20 cm from the table top has been taken as a threshold, T value). The same height has been maintained till the gripper reaches to the targeted object. 
    \item Subsequently, the robot gripper is allowed to come down by manipulating the movement in the negative z direction by an amount of $\lvert z-(depth(gpc)+sd) \rvert$ from the table top where gpc is the grasp patch center and sd is a safe distance.
    Also, depending on the height of the object to be grasped, sometimes it is a good idea to provide a safe distance factor which is the $20\%(height(gpc))$ from the table top, 
    to avoid gripper hitting on the table top. In our experiments we have frequently used this safe distance factor.
    \item Finally, the gripper moves at a distance/height where $z=0$ on the table top, which is defined to be its new home position to complete  the execution cycle. 
\end{itemize}

\noindent
\begin{figure}[!t]
        \begin{subfigure}[b]{0.25\textwidth}
                \centering
                \includegraphics[width=0.65\linewidth]{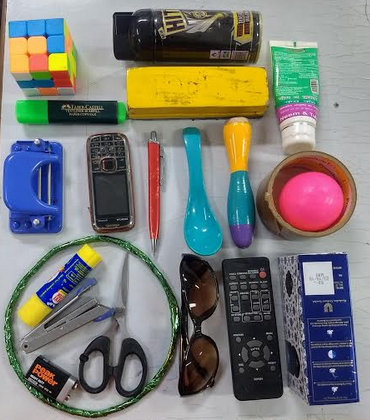}
                \caption{Seen}
	    \label{fig:objectseen}
        \end{subfigure}%
        \begin{subfigure}[b]{0.25\textwidth}
                \centering
                \includegraphics[width=0.60\linewidth]{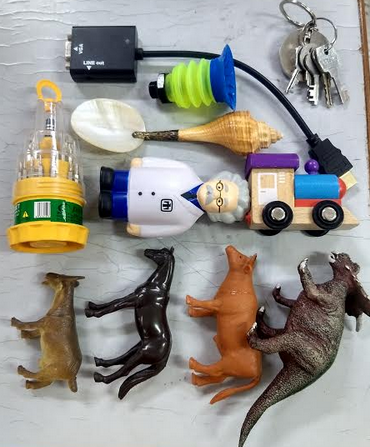}
                \caption{Unseen}
	    \label{fig:objectunseen}
        \end{subfigure}
        
        \caption{Experimental objects}\label{fig:ges}
\end{figure}

\section{Experiments with the robot}
To demonstrate the experimental part of our proposed approach, this section has been subdivided into three subsections. First building the Robotic experimental setup has been described with hardware and software configurations followed by the description of experimental details using training and testing samples which include unseen/novel objects during testing. In the subsequent sections, we analysed the results obtained during training and testing.
 
\begin{figure}[!t]
	\centering
		\includegraphics[scale=0.4]{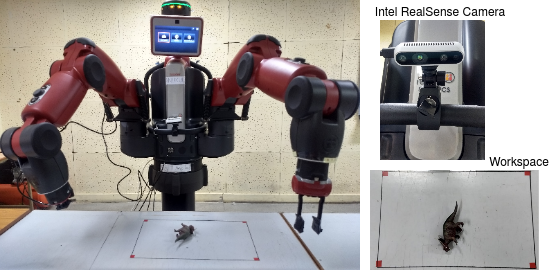}
	\caption{Experimental Setup}
	\label{fig:setup}
\end{figure}

\subsection{Description of Robotic experimental setup}
For vision based grasping, we have used external camera, Intel ``RealSense" having model no. D435. It uses stereo-vision for depth estimation  and consists of the paired RGB sensors together with depth sensors. For manipulation task Anukul research robot, available at our lab, has been used. It has two arms with seven degrees of freedom. To map the robot arm with camera, left arm has been considered during experimentation which can be replaced with right arm without affecting performance. For viewing the working space camera has been mounted over the torso of the Anukul research robot. The camera saves the image with $x\times y$ dimensions, then it has been cropped as per specified red corner mark on the table workspace. 
 The complete robotic experimental setup  has been shown in Fig. \ref{fig:setup}. For running all the robot experiments Baxter SDK workstation setup has been used along with the kinetic version of the Robot Operating System (ROS) on top of the Ubuntu 16.04 platform.

\subsection{Description of the Experiments}
To evaluate the proposed approach, first we have trained our VQVAE and GGCNN2 based RGGCNN2 model with Cornell dataset which is ordinarily a labelled dataset. However, we have divided this dataset into labelled (n1) and unlabelled (n2) datasets with different ratios such as 0.1, 0.3, 0.5, 0.7 and 0.9. Ratio has been calculated as n1:(n1+n2) where n1 defines the fraction of labelled data which have been used for supervised grasp patch learning. 
For model training Central Computing Facility (CCF) has been used which provides GPU queue for each user with configuration, such as 40 cores and 384 GB  RAM with CUDA core 10.240 in each NVIDIA Tesla-V100 cluster. 
The dataset consists of 885 images with 2209 positive and 5116 negative  grasp patch/rectangle annotations. 

To evaluate the performance of the  trained model in real time environment, different categories of objects have been used. During  experiments with robot, objects have been divided into two groups such as seen and unseen. Both the groups include regular and irregular shaped objects which have been shown in  Fig. \ref{fig:objectseen} and \ref{fig:objectunseen}. Each object has been tested for 10 different positions, with  4 different orientations. For seen object category, 20 objects have been used, whereas 10 objects have been used for unseen category. Throughout our experiments, it is observed that proposed approaches are largely effective for seen as well as unseen objects. Due to table top manipulation, we have considered small to medium sized objects which are feasible to be grasped by the robotic arm along with electrically driven gripper. However, it should also work efficiently with the vacuum gripper for regular shaped objects. 




\begin{figure*}
\centering
\begin{adjustbox}{width=\textwidth}
\begin{tikzpicture}
    \begin{axis}[
    width=0.9\textwidth,
    height=3cm,
    ymajorgrids,
    ylabel=$Test$ $Accuracy$ $(\%)$,
    xlabel=$Labelled$ $Data$ $Ratio$,
    legend pos=outer north east
    ]
        \addplot table[x=in,y=rggcnn2] {plotData.csv};\addlegendentry{RGGCNN2 (Ours)}
        \addplot table[x=in,y=rggcnn] {plotData.csv};\addlegendentry{RGGCNN}
        
    \end{axis}
\end{tikzpicture}
\end{adjustbox}
\caption{Performance comparison between RGGCNN and RGGCNN2 models.} \label{fig:RGGCNN2RGGCNN}
\end{figure*}
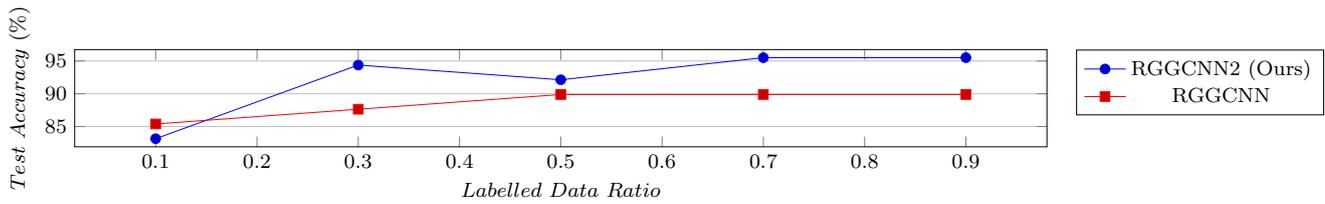

\begin{table*}
\caption{Trained model's performance evaluation}
\label{table:iou}
\centering
\begin{tabular}{|c|c|c|c|c|}
\hline
\textbf{Ratio} & \textbf{RGGCNN} \cite{mridul} & \textbf{RGGCNN2} (Ours)\\
\hline
 \textbf{0.1} &  85.3933 & 83.1461\\ 
 \hline
 \textbf{0.3} &  87.6404 & 94.3820\\
 \hline
 \textbf{0.5} &  89.8876 & 92.1348\\
 \hline
 \textbf{0.7} &  89.8876 & 95.5056\\
 \hline
 \textbf{0.9} &  89.8876 & 95.5056\\ 
\hline
\end{tabular}
\end{table*}

\begin{table}[!ht]
\centering
 \begin{tabular}{| c | c |} 
 \hline
 \textbf{Model} & \textbf{Accuracy (\%)}  \\ 
 \hline
 GGCNN \cite{morrison2018closing} & 73.0  \\\hline
 SAE, struct. reg. \cite{lenz2015deep} & 73.9  \\\hline
 Two-stage closed-loop \cite{wang2016robot} & 85.3  \\\hline
 AlexNet, MultiGrasp \cite{redmon2015real} & 88.0  \\\hline 
 STEM-CaRFs \cite{asif2018ensemblenet} & 88.2  \\\hline 
 GRPN \cite{karaoguz2019object} & 88.7  \\\hline 
 ResNet-50x2 \cite{kumra2017robotic} & 89.2  \\\hline
 RGGCNN \cite{mridul} & 89.8  \\\hline 
 GraspNet \cite{asif2018graspnet} & 90.2 \\\hline 
 ZF-net \cite{guo2017hybrid} & 93.2  \\\hline 
 \textbf{RGGCNN2 (Ours)} & \textbf{95.5}  \\\hline 
 \end{tabular}
 \caption{Performance comparison on Cornell dataset}
 \label{table:Tab2}
\end{table}

\begin{figure*}
    \minipage{0.48\textwidth}
      \includegraphics[width=0.95\columnwidth]{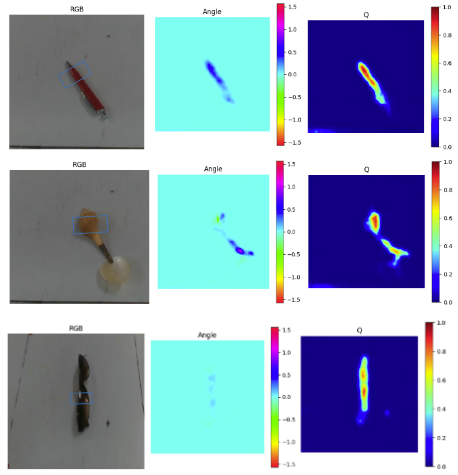}
    \endminipage
    \minipage{0.48\textwidth}
      \includegraphics[width=0.95\columnwidth]{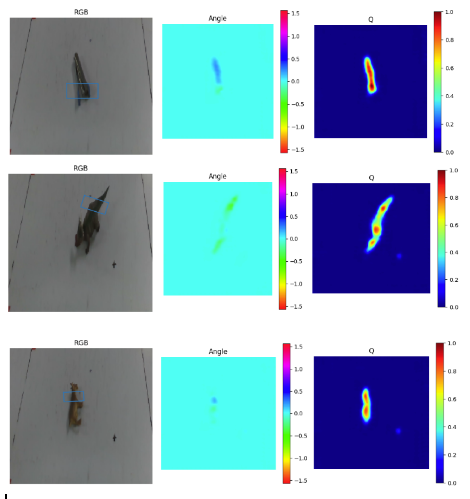}
    \endminipage
\caption{RGGCNN2 results with seen and unseen objects}
\label{fig:isolatedresult}
\end{figure*}

\begin{figure*}[t]
	\centering
    \includegraphics[scale=1.2]{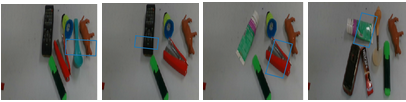}
    \caption{RGGCNN2 results within cluttered environment}
    \label{fig:clutteredresult}
\end{figure*}

\subsection{Result analyses}
Primarily, the performance of the trained models have been evaluated with varying labelled and unlabelled data set ratios. More specifically, during training we wanted to test our model about how much generalization it can achieve with unlabelled data. In the semi supervised learning paradigm, if we can use more and more unlabelled data for our proposed model, RGGCNN2, it will be desirable and hence it's  performance has been compared with the baseline architecture as proposed in \cite{mridul,priya}, which is  termed as RGGCNN here, with different blend of labelled and unlabelled data sets. The preliminary idea being appreciated in CVPR, Women in Computer Vision Workshop (Poster) (2020) and SPCOM(2020) \cite{mridul,priya}, encouraged us to pursue full blown research in this area. Both the models have been tested for five different ratios and their performance, in terms of Intersection Over Union (IOU), have been presented in Fig. \ref{fig:RGGCNN2RGGCNN}. From the Fig. \ref{fig:RGGCNN2RGGCNN}, it can be infer that the RGGCNN2 outperforms RGGCNN when it has only sufficient labelled dataset to train a grasp prediction model. All the evaluated model's performance have been tabulated in Table \ref{table:iou} for clear visibility. The reported accuracy of RGGCNN is increased by 6\% (approx.) in our proposed RGGCNN2.
The performance of RGGCNN2 has been also compared with the state-of-the-art models as tabulated in Table \ref{table:Tab2}. During experiments, performance of our proposed model has been validated based on image-wise split to test the generalization capability of the model.

Further, the performance of our proposed trained model, RGGCNN2, have been tested with both seen and unseen objects. The predicted grasp results have been represented by a Quality map (Q map) and Angle map for each object where Q map and angle map uses color gradients to represent the numerical values within a defined range. During the  validation of the model, it has been observed that our approach works significantly better in representing the color gradient values with the normal as well as dim lights. Due to the inherent strength of our proposed model of getting trained with both unsupervised (with unlabelled data for learning embedding by the vector quantizer) and supervised (with labelled data for grasp patch learning) way (which we are calling a semi supervised learning), it can predict optimal grasp patches for seen as well as  unseen objects, due to better generalization which are visible from the results as shown in Fig. \ref{fig:isolatedresult} and Fig. \ref{fig:clutteredresult}. 
In the summary, we can say that our proposed model can be very effective when we really do not have enough labelled data, sufficient for training a giant network like this, and can efficiently predict grasp patches both for seen and unseen objects and in an isolated as well as cluttered environments and can act as a bench mark procedure to train such networks.  

\section{Conclusion and Future work}
In the present investigation, we have addressed the problem of solving intelligent object grasping using lesser number of labelled data. It has been demonstrated that the  representation based grasp patch detection can help finding correct grasp patches using the semi supervised learning technique. Our proposed approach, RGGCNN2, uses RGB images to generate grasp patch in the isolated as well as cluttered environment. By virtue of using VQVAE for semi supervised learning, our model could achieve the ability to predict and grasp not only the previously seen objects used for training, but also similar objects which robot has not seen so far. From the results (refer Table \ref{table:iou} and Table \ref{table:Tab2}), it can be seen that our RGGCNN2 model works significantly better in comparison to the other existing state of the art models. Moreover, predicted grasp patch for an object has been also evaluated with the physical robot. 

Some of the limitations, which can be taken up for future research, include improper object grasp patch prediction due to imprecise depth estimation for reflective and transparent objects. Grasp patch prediction failure can also occur due to background colour similarity with the object or presence of multiple objects having similar colour. 
Future, research can also be directed towards refining the latent space for better posterior and to extend the model's capability by including touch sensors which can include the grasp affordances, so that the robot can grasp soft and fragile objects.

\bibliographystyle{spmpsci}
\bibliography{bib1}

\begin{thebibliography}{10}
\providecommand{\url}[1]{{#1}}
\providecommand{\urlprefix}{URL }
\expandafter\ifx\csname urlstyle\endcsname\relax
  \providecommand{\doi}[1]{DOI~\discretionary{}{}{}#1}\else
  \providecommand{\doi}{DOI~\discretionary{}{}{}\begingroup
  \urlstyle{rm}\Url}\fi

\bibitem{asif2018ensemblenet}
Asif, U., Tang, J., Harrer, S.: Ensemblenet: Improving grasp detection using an
  ensemble of convolutional neural networks.
\newblock In: BMVC, p.~10 (2018)

\bibitem{asif2018graspnet}
Asif, U., Tang, J., Harrer, S.: Graspnet: An efficient convolutional neural
  network for real-time grasp detection for low-powered devices.
\newblock In: IJCAI, vol.~7, pp. 4875--4882 (2018)

\bibitem{c3}
{Bohg}, J., {Morales}, A., {Asfour}, T., {Kragic}, D.: Data-driven grasp
  synthesis—a survey.
\newblock IEEE Transactions on Robotics \textbf{30}(2), 289--309 (2014)

\bibitem{jacquard}
Depierre, A., Dellandr{\'{e}}a, E., Chen, L.: Jacquard: {A} large scale dataset
  for robotic grasp detection.
\newblock CoRR \textbf{abs/1803.11469} (2018).
\newblock \urlprefix\url{http://arxiv.org/abs/1803.11469}

\bibitem{ggcnn2}
{Douglas Morrison and Peter Corke and J{\"{u}}rgen Leitner}: Learning robust,
  real-time, reactive robotic grasping.
\newblock The International Journal of Robotics Research \textbf{39}(2-3),
  183--201 (2020).
\newblock \doi{10.1177/0278364919859066}.
\newblock \urlprefix\url{https://doi.org/10.1177/0278364919859066}

\bibitem{n11}
Finn, C., Goodfellow, I.J., Levine, S.: Unsupervised learning for physical
  interaction through video prediction.
\newblock CoRR \textbf{abs/1605.07157} (2016).
\newblock \urlprefix\url{http://arxiv.org/abs/1605.07157}

\bibitem{n12}
Goodfellow, I.J., Pouget-Abadie, J., Mirza, M., Xu, B., Warde-Farley, D.,
  Ozair, S., Courville, A., Bengio, Y.: Generative adversarial nets.
\newblock In: Proceedings of the 27th International Conference on Neural
  Information Processing Systems - Volume 2, NIPS'14, p. 2672–2680. MIT
  Press, Cambridge, MA, USA (2014)

\bibitem{guo2017hybrid}
Guo, D., Sun, F., Liu, H., Kong, T., Fang, B., Xi, N.: A hybrid deep
  architecture for robotic grasp detection.
\newblock In: 2017 IEEE International Conference on Robotics and Automation
  (ICRA), pp. 1609--1614. IEEE (2017)

\bibitem{KL}
{Ji}, S., {Zhang}, Z., {Ying}, S., {Wang}, L., {Zhao}, X., {Gao}, Y.:
  Kullback-leibler divergence metric learning.
\newblock IEEE Transactions on Cybernetics pp. 1--12 (2020).
\newblock \doi{10.1109/TCYB.2020.3008248}

\bibitem{BK}
{Ju}, Z., {Yang}, C., {Ma}, H.: Kinematics modeling and experimental
  verification of baxter robot.
\newblock In: Proceedings of the 33rd Chinese Control Conference, pp.
  8518--8523 (2014).
\newblock \doi{10.1109/ChiCC.2014.6896430}

\bibitem{n20}
Kalchbrenner, N., van~den Oord, A., Simonyan, K., Danihelka, I., Vinyals, O.,
  Graves, A., Kavukcuoglu, K.: Video pixel networks.
\newblock CoRR \textbf{abs/1610.00527} (2016).
\newblock \urlprefix\url{http://arxiv.org/abs/1610.00527}

\bibitem{karaoguz2019object}
Karaoguz, H., Jensfelt, P.: Object detection approach for robot grasp
  detection.
\newblock In: 2019 International Conference on Robotics and Automation (ICRA),
  pp. 4953--4959. IEEE (2019)

\bibitem{RPT}
Kingma, D.P., Welling, M.: Auto-encoding variational bayes (2014)

\bibitem{p9}
Konidaris, G., Kuindersma, S., Grupen, R., Barto, A.: Robot learning from
  demonstration by constructing skill trees.
\newblock The International Journal of Robotics Research \textbf{31}(3),
  360--375 (2012).
\newblock \doi{10.1177/0278364911428653}.
\newblock \urlprefix\url{https://doi.org/10.1177/0278364911428653}

\bibitem{agc}
Ku, L.Y., Learned{-}Miller, E.G., Grupen, R.A.: Associating grasping with
  convolutional neural network features.
\newblock CoRR \textbf{abs/1609.03947} (2016).
\newblock \urlprefix\url{http://arxiv.org/abs/1609.03947}

\bibitem{kumra2017robotic}
Kumra, S., Kanan, C.: Robotic grasp detection using deep convolutional neural
  networks.
\newblock 2017 IEEE/RSJ International Conference on Intelligent Robots and
  Systems (IROS) pp. 769--776 (2017)

\bibitem{lenz2015deep}
Lenz, I., Lee, H., Saxena, A.: Deep learning for detecting robotic grasps.
\newblock The International Journal of Robotics Research \textbf{34}(4-5),
  705--724 (2015)

\bibitem{d1}
Levine, S., Pastor, P., Krizhevsky, A., Quillen, D.: Learning hand-eye
  coordination for robotic grasping with deep learning and large-scale data
  collection.
\newblock CoRR \textbf{abs/1603.02199} (2016).
\newblock \urlprefix\url{http://arxiv.org/abs/1603.02199}

\bibitem{od2}
Liu, W., Anguelov, D., Erhan, D., Szegedy, C., Reed, S.E., Fu, C., Berg, A.C.:
  {SSD:} single shot multibox detector.
\newblock CoRR \textbf{abs/1512.02325} (2015).
\newblock \urlprefix\url{http://arxiv.org/abs/1512.02325}

\bibitem{mridul}
{Mahajan}, M., {Bhattacharjee}, T., {Krishnan}, A., {Shukla}, P., {Nandi},
  G.C.: Robotic grasp detection by learning representation in a vector
  quantized manifold.
\newblock In: 2020 International Conference on Signal Processing and
  Communications (SPCOM), pp. 1--5 (2020)

\bibitem{c21}
Mahler, J., Liang, J., Niyaz, S., Laskey, M., Doan, R., Liu, X., Ojea, J.A.,
  Goldberg, K.: Dex-net 2.0: Deep learning to plan robust grasps with synthetic
  point clouds and analytic grasp metrics.
\newblock CoRR \textbf{abs/1703.09312} (2017).
\newblock \urlprefix\url{http://arxiv.org/abs/1703.09312}

\bibitem{he}
{Maitin-Shepard}, J., {Cusumano-Towner}, M., {Lei}, J., {Abbeel}, P.: Cloth
  grasp point detection based on multiple-view geometric cues with application
  to robotic towel folding.
\newblock In: 2010 IEEE International Conference on Robotics and Automation,
  pp. 2308--2315 (2010)

\bibitem{n26}
Mehri, S., Kumar, K., Gulrajani, I., Kumar, R., Jain, S., Sotelo, J.,
  Courville, A.C., Bengio, Y.: Samplernn: An unconditional end-to-end neural
  audio generation model.
\newblock CoRR \textbf{abs/1612.07837} (2016).
\newblock \urlprefix\url{http://arxiv.org/abs/1612.07837}

\bibitem{morrison2018closing}
Morrison, D., Corke, P., Leitner, J.: Closing the loop for robotic grasping:
  {A} real-time, generative grasp synthesis approach.
\newblock CoRR \textbf{abs/1804.05172} (2018).
\newblock \urlprefix\url{http://arxiv.org/abs/1804.05172}

\bibitem{n37}
van~den Oord, A., Dieleman, S., Zen, H., Simonyan, K., Vinyals, O., Graves, A.,
  Kalchbrenner, N., Senior, A.W., Kavukcuoglu, K.: Wavenet: {A} generative
  model for raw audio.
\newblock CoRR \textbf{abs/1609.03499} (2016).
\newblock \urlprefix\url{http://arxiv.org/abs/1609.03499}

\bibitem{n38}
van~den Oord, A., Kalchbrenner, N., Vinyals, O., Espeholt, L., Graves, A.,
  Kavukcuoglu, K.: Conditional image generation with pixelcnn decoders.
\newblock CoRR \textbf{abs/1606.05328} (2016).
\newblock \urlprefix\url{http://arxiv.org/abs/1606.05328}

\bibitem{VQVAE}
van~den Oord, A., Vinyals, O., Kavukcuoglu, K.: Neural discrete representation
  learning.
\newblock CoRR \textbf{abs/1711.00937} (2017).
\newblock \urlprefix\url{http://arxiv.org/abs/1711.00937}

\bibitem{c23}
Pinto, L., Gupta, A.: Supersizing self-supervision: Learning to grasp from 50k
  tries and 700 robot hours.
\newblock CoRR \textbf{abs/1509.06825} (2015).
\newblock \urlprefix\url{http://arxiv.org/abs/1509.06825}

\bibitem{redmon2015real}
Redmon, J., Angelova, A.: Real-time grasp detection using convolutional neural
  networks.
\newblock In: 2015 IEEE International Conference on Robotics and Automation
  (ICRA), pp. 1316--1322. IEEE (2015)

\bibitem{od1}
Redmon, J., Divvala, S.K., Girshick, R.B., Farhadi, A.: You only look once:
  Unified, real-time object detection.
\newblock CoRR \textbf{abs/1506.02640} (2015).
\newblock \urlprefix\url{http://arxiv.org/abs/1506.02640}

\bibitem{c27}
Sahbani, A., El-Khoury, S., Bidaud, P.: An overview of 3d object grasp
  synthesis algorithms.
\newblock Robotics and Autonomous Systems \textbf{60}, 326--336 (2012).
\newblock \doi{10.1016/j.robot.2011.07.016}

\bibitem{13}
Schmidt, P., Vahrenkamp, N., W{\"a}chter, M., Asfour, T.: Grasping of unknown
  objects using deep convolutional neural networks based on depth images.
\newblock 2018 IEEE International Conference on Robotics and Automation (ICRA)
  pp. 6831--6838 (2018)

\bibitem{priya}
Shukla, P., Mahajan, M., Bhattacharjee, T., Krishnan, A., Nandi, G.C.: Robotic
  grasp detection by learning representation in a vector quantized manifold.
\newblock In: CVPR, Women in Computer Vision Workshop (Poster) (2020)

\bibitem{PD}
{Tchuiev}, V., {Indelman}, V.: Inference over distribution of posterior class
  probabilities for reliable bayesian classification and object-level
  perception.
\newblock IEEE Robotics and Automation Letters \textbf{3}(4), 4329--4336
  (2018).
\newblock \doi{10.1109/LRA.2018.2852844}

\bibitem{cgd}
University, C.: Robot learning lab: Learning to grasp.
\newblock Available online: \url{http://pr.cs.cornell.edu/
  grasping/rect\_data/data.php}

\bibitem{d11}
Viereck, U., ten Pas, A., Saenko, K., Jr., R.P.: Learning a visuomotor
  controller for real world robotic grasping using easily simulated depth
  images.
\newblock CoRR \textbf{abs/1706.04652} (2017).
\newblock \urlprefix\url{http://arxiv.org/abs/1706.04652}

\bibitem{st}
Vinyals, O., Toshev, A., Bengio, S., Erhan, D.: Show and tell: {A} neural image
  caption generator.
\newblock CoRR \textbf{abs/1411.4555} (2014).
\newblock \urlprefix\url{http://arxiv.org/abs/1411.4555}

\bibitem{wang1999lagrangian}
Wang, J., Hu, Q., Jiang, D.: A lagrangian network for kinematic control of
  redundant robot manipulators.
\newblock IEEE Transactions on Neural Networks \textbf{10}(5), 1123--1132
  (1999)

\bibitem{wang2016robot}
Wang, Z., Li, Z., Wang, B., Liu, H.: Robot grasp detection using multimodal
  deep convolutional neural networks.
\newblock Advances in Mechanical Engineering \textbf{8}(9), 1687814016668077
  (2016)

\bibitem{wei2017robotic}
Wei, J., Liu, H., Yan, G., Sun, F.: Robotic grasping recognition using
  multi-modal deep extreme learning machine.
\newblock Multidimensional Systems and Signal Processing \textbf{28}(3),
  817--833 (2017)

\bibitem{14}
Zeng, A., Song, S., Yu, K., Donlon, E., Hogan, F.R., Bauz{\'{a}}, M., Ma, D.,
  Taylor, O., Liu, M., Romo, E., Fazeli, N., Alet, F., Dafle, N.C., Holladay,
  R., Morona, I., Nair, P.Q., Green, D., Taylor, I.J., Liu, W., Funkhouser,
  T.A., Rodriguez, A.: Robotic pick-and-place of novel objects in clutter with
  multi-affordance grasping and cross-domain image matching.
\newblock CoRR \textbf{abs/1710.01330} (2017).
\newblock \urlprefix\url{http://arxiv.org/abs/1710.01330}

\end{thebibliography}
\end{document}